\documentclass[nonacm,sigconf,natbib=false]{acmart}

\usepackage{amsmath, amsfonts}
\usepackage{enumitem}
\usepackage{graphicx}
\usepackage{indentfirst}
\usepackage[backend=biber, style=numeric]{biblatex}
\usepackage{multirow}
\usepackage[subpreambles=true]{standalone}
\usepackage{todonotes}

\renewcommand\vec[1]{\boldsymbol{#1}}

\title{Alloprof: a French question-answer education dataset and its use in an information retrieval case study}
\date{}

\author{Antoine Lefebvre-Brossard}
\email{antoine.lefebvre-brossard@polymtl.ca}
\affiliation{
  \institution{Mila - Quebec AI Institute}
  \country{}
}
\affiliation{
  \institution{Polytechnique Montréal}
  \country{}
}
\author{Stephane Gazaille}
\email{stephane.gazaille@mila.quebec}
\affiliation{
  \institution{Mila - Quebec AI Institute}
  \country{}
}
\author{Michel C. Desmarais}
\email{michel.desmarais@polymtl.ca}
\affiliation{
  \institution{Polytechnique Montréal}
  \country{}
}

\begin{document}

\sloppy

\begin{abstract}
  \textit{Context:}
  Teachers and students are increasingly relying on online learning resources to supplement the ones provided in school.
  This increase in the breadth and depth of available resources is a great thing for students, but only provided they are able to find answers to their queries.
  Question-answering and information retrieval systems alleviate the task of finding the relevant resources and public datasets of labeled data are essential to train and evaluate their algorithms, but most of these datasets are English text written by and for adults.

  \textit{Objectives:}
  We introduce a new public French question-answering dataset collected from Alloprof, a Quebec-based primary and high-school help website, containing 29~349 questions and their explanations in a variety of school subjects from 10~368 students, with more than half of the explanations containing links to other questions or some of the 2~596 reference pages on the website.
  We also present a case study using this dataset for an information retrieval task.

  \textit{Method:}
  This dataset was collected on the Alloprof public forum, with all questions verified for their appropriateness and the explanations verified both for their appropriateness and their relevance to the question.
  To predict relevant documents (questions or reference pages), architectures using pre-trained BERT models were fine-tuned and evaluated on their ability to predict at least one document referred to in the explanation from their top-3 predictions for a question, as well as on the MRR and nDCG metrics.

  \textit{Results:}
  The best model obtains a prediction score of 58.5\% (MRR: 0.54, nDCG: 0.62).  These results are substantially better than a TF-IDF vector space model (prediction score: 0.24, MRR: 0.20, nDCG: 0.32), but their computational cost is also greater. The fastest model obtains a score of 51.2\% (MRR:0.467, nDCG: 0.560) with an inference time of 0.7s, which is still considered acceptable in the context of Alloprof.

  \textit{Conclusions:}
  This dataset will allow researchers to develop question-answering, information retrieval and other algorithms specifically for the French speaking education context.
  Furthermore, the range of language proficiency, images, mathematical symbols and spelling mistakes will necessitate algorithms based on a multimodal comprehension.
  The case study we present as a baseline shows an approach that relies on recent techniques provides an acceptable performance level, but more work is necessary before it can reliably be used and trusted in a production setting.
\end{abstract}


\maketitle

\section{Introduction}

While the tendency to turn to the Internet to find learning material is well established, we find a growing tendency from students and teachers alike to request, submit, and exchange such material.
This phenomenon is nourished in part by the social aspects that some learning environments provide, and by the increased ease at which one can find or submit learning material.  
In Quebec, a well-known and used website established by the non-profit Alloprof\footnote{\url{https://www.alloprof.qc.ca}} has been providing access for years already to well-crafted reference pages on most concepts primary and high-school students would encounter in their schooling as well as a public forum where students could ask questions which would be answered by other users, whether these be students or teachers, and verified as accurate or not by moderators.

To help the goal of developing and validating means to match queries from students with the appropriate learning material, we introduce a French question-answering dataset in education created through a partnership between Alloprof and Mila.
Covering a six months period, this dataset contains 29~349 questions and their explanation in a variety of school subjects from 10~368 students, with more than half of the explanations containing links to other questions or some of the 2~596 reference pages on the website.
We make available this dataset in the original raw format and a processed CSV file on GitHhub\footnote{\url{https://github.com/mila-aia/alloprof-data}} and HuggingFace\footnote{\url{https://huggingface.co/datasets/antoinelb7/alloprof}}.
The code to produce the CSV file from the raw data is also available in the GitHub repo.

We use this dataset in a case study where we create an information retrieval system predicting the most relevant old questions or reference pages for a given student question in both a precise and fast manner.
This system is to be used on the Alloprof website to help as many students find the relevant learning material as quickly as possible so they can stay focused on their studies and accomplish them more quickly, rather than have to wait for someone to answer their question.
\footnote{Please contact the first author to gain access to the code used to train and serve the models.}

We explain how this dataset was created and its characteristics in section~\ref{sec:dataset}, and in section~\ref{sec:case-study}, describe its use in the information retrieval case study.
  
\subsection{Related Work}

In recent years, there has been a multitude of question-answering datasets released to serve as benchmarks for new models~\cite{lai17,raj16,raj18,red18}, with some even being in French~\cite{hof20,hei22,kad22,ker20}.
While most of these were collected from Wikipedia articles, \cite{lai17} and \cite{red18} are notable in the context of this paper for also including middle and high school English exam questions.

Information retrieval also has a long history of public datasets used to compare information retrieval systems~\cite{har93}, most notably the TREC datasets~\cite{sob21}, provided since 1993 to evaluate information retrieval in a standardised way on large-scale evaluations of text retrieval methodologies, and in more recent years, MS MARCO~\cite{ngu16}, a dataset initially used for question answering. These datasets are in English, but others in different languages are starting to be released such as BSARD~\cite{lou22} and MIRACL~\cite{zha22}.

These datasets differ from the one introduced here in two main regards.
While these question-answering and information retrieval datasets are overwhelmingly textual, this one has many queries containing images.
The textual complexity will also be different as public information retrieval contain queries and documents written by adults whereas this dataset is composed of queries written by children and labeled by their school grade, roughly 1 through 11~(see section~\ref{sec:characteristics}).
This means that a great question-answering or information retrieval system for this dataset would need to not only be able to understand queries written in a large variety of language and thought sophistication, but also multimodal ones, combining the textual, symbolic and visual information.

To solve the information retrieval task, recent work has shown the advantage of using transformer-based approaches~\cite{hof21,nog19,tra21} over more classical probabilistic bag-of-words based ones~\cite{rob09}.
These are usually separated in representation-based models, which encode both the query and document in a vector space and then use a similarity metric to select the best matching documents, and interaction-based models, which will use both the query and document to predict a relevance using the interaction between the terms from both of them~\cite{abb20}.

\section{Dataset}
\label{sec:dataset}

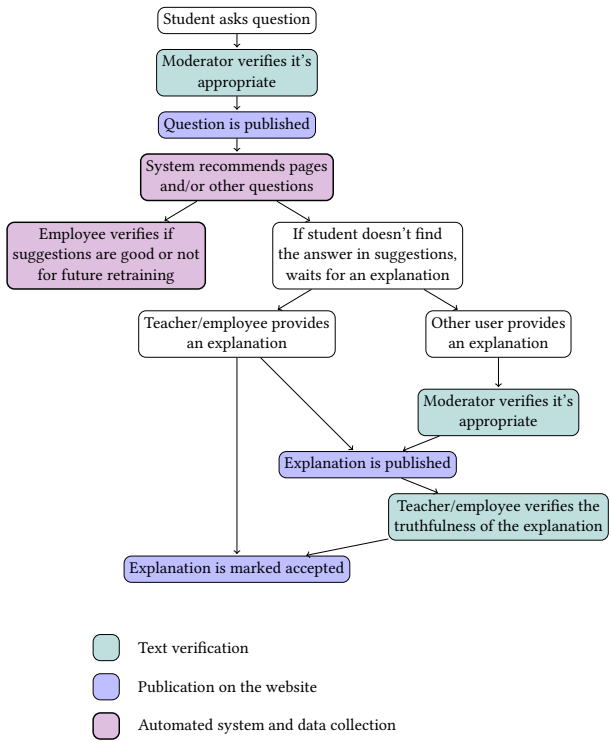
\begin{figure}[ht]
  \begin{center}
    \begin{tikzpicture}[
  dot/.style = {circle, fill, minimum size=#1,
  inner sep=0pt, outer sep=0pt},
  dot/.default = 6pt
  ]  

  \node[anchor=center, align=center, rectangle, rounded corners, draw]
    (s) at (0, 0)
    {Student asks question};
  \node[anchor=center, align=center, rectangle, rounded corners, draw, fill=teal!25]
    (m1) at (0, -1)
    {Moderator verifies it's \\ appropriate};
  \node[anchor=center, align=center, rectangle, rounded corners, draw, fill=blue!25]
    (q) at (0, -2)
    {Question is published};
  \node[anchor=center, align=center, rectangle, rounded corners, draw, fill=violet!25, thick]
    (r) at (0, -3)
    {System recommends pages \\ and/or other questions};
  \node[anchor=center, align=center, rectangle, rounded corners, draw, fill=violet!25, thick]
    (v) at (-2.5, -4.5)
    {Employee verifies if \\ suggestions are good or not \\ for future retraining};
  \node[anchor=center, align=center, rectangle, rounded corners, draw]
    (w) at (2.5, -4.5)
    {If student doesn't find \\ the answer in suggestions, \\ waits for an explanation};
  \node[anchor=center, align=center, rectangle, rounded corners, draw]
    (t) at (0, -6)
    {Teacher/employee provides \\ an explanation};
  \node[anchor=center, align=center, rectangle, rounded corners, draw]
    (o) at (5, -6)
    {Other user provides \\ an explanation};
  \node[anchor=center, align=center, rectangle, rounded corners, draw, fill=teal!25]
    (m2) at (5, -7.5)
    {Moderator verifies it's \\ appropriate};
  \node[anchor=center, align=center, rectangle, rounded corners, draw, fill=blue!25]
    (p) at (2.5, -8.5)
    {Explanation is published};
  \node[anchor=center, align=center, rectangle, rounded corners, draw, fill=teal!25]
    (tr) at (5, -9.5)
    {Teacher/employee verifies the \\ truthfulness of the explanation};
  \node[anchor=center, align=center, rectangle, rounded corners, draw, fill=blue!25]
    (a) at (0, -10.5)
    {Explanation is marked accepted};

  \node[
    anchor=center,
    align=center,
    rectangle,
    rounded corners,
    draw,
    fill=teal!25,
    minimum width=0.5cm,
    minimum height=0.5cm,
    ] at (-2.5, -12) {};
  \node[anchor=west, align=left]
    at (-2, -12)
    {Text verification};
  \node[
    anchor=center,
    align=center,
    rectangle,
    rounded corners,
    draw,
    fill=blue!25,
    minimum width=0.5cm,
    minimum height=0.5cm,
    ] at (-2.5, -12.75) {};
  \node[anchor=west, align=left]
    at (-2, -12.75)
    {Publication on the website};
  \node[
    anchor=center,
    align=center,
    rectangle,
    rounded corners,
    draw,
    fill=violet!25,
    thick,
    minimum width=0.5cm,
    minimum height=0.5cm,
    ] at (-2.5, -13.5) {};
  \node[anchor=west, align=left]
    at (-2, -13.5)
    {Automated system and data collection};

    \draw[->, shorten >=1pt] (s) -- (m1);
  \draw[->, shorten >=1pt] (m1) -- (q);
  \draw[->, shorten >=1pt] (q) -- (r);
  \draw[->, shorten >=1pt] (r) -- (v);
  \draw[->, shorten >=1pt] (r) -- (w);
  \draw[->, shorten >=1pt] (w) -- (t);
  \draw[->, shorten >=1pt] (w) -- (o);
  \draw[->, shorten >=1pt] (o) -- (m2);
  \draw[->, shorten >=1pt] (t) -- (p);
  \draw[->, shorten >=1pt] (m2) -- (p);
  \draw[->, shorten >=1pt] (t) -- (a);
  \draw[->, shorten >=1pt] (p) -- (tr);
  \draw[->, shorten >=1pt] (tr) -- (a);

\end{tikzpicture}
  \end{center}
  \caption{The current system currently used at Alloprof, as well as the modifications to be put in place to use the information retrieval system and collect more data. Every step is verified to make sure only appropriate questions and explanations are published, as well as make sure correct explanations are well identified.}
  \label{fig:alloprof}
\end{figure}

The data comes from a public forum\footnote{\url{https://www.alloprof.qc.ca/helpzone/discussions}} on the Alloprof site where students ask questions which can be answered by teachers or any other registered user.
As figure~\ref{fig:alloprof} shows, verification is done at each step by moderators or teachers to validate that only appropriate and truthful content is published.

When asking a question, a student will mostly write text, but can also add images to give context to their question, and will identify the grade they are in as well as the school subject corresponding to the question.
The text will often contain spelling or grammar mistakes, and may contain formulas or links, while the images can be a screenshot of a printed question they are referring to, a picture of a handwritten answer to an exam they would like explained, a mathematical diagram or formula, or anything else related to their question.
A moderator verifies the question is appropriate and approves its publication, but does not make any assessment about the possibility of being able to answer the question, whether that be because it is too convoluted to make sense or the image is not legible enough.
This process ensures that every question in the dataset is a genuine one from a student and relevant to their education.

An explanation can come from two different sources.
In the first case, a teacher or Alloprof employee will give an explanation to a question and it will be automatically published and marked as accepted, which means that the explanation can be considered a correct one for the question.
In the second case, another user of the platform can write an explanation for the question.
In that case, a moderator verifies that the explanation is appropriate (but not necessarily the right one) before it is published on the site.
Then a teacher or employee must verify that the explanation is right so it can be marked as accepted.
A question can have multiple explanations, but only one will be marked as accepted.
Like for a question, an explanation will mostly contain text, but may also contain images.
But one thing that differentiates them is the common use of links to Alloprof reference pages or other questions containing an explanation to the question.

We use the term \textit{explanation} rather than \textit{answer} to emphasize the fact that the goal of the forum is not to give the exact answer to the question, but to help the student understand and find their own answer by guiding them with that explanation.

Table~\ref{tab:example} shows an example of a question with its corresponding answer containing a link to a reference page.

\begin{table}
  \begin{tabular}{| p{0.2\linewidth} p{0.7\linewidth} |}
    \hline
    Language: & en \\
    Subject: & financial\_ed \\
    Grade: & Secondaire 5 (12) \\
    Question: & Hello, Can you help me find the consequences of tourism on the Maritime Greenwich in London. I've been looking for 3 hours. Thank you \\
    Explanation: & Hello :) On the following site, you will find several aspects threatening the conservation of the Maritime Greenwich near London. I suggest you start your reading from the "Integrity" section: https://whc.unesco.org/en/list/795/ Finally, I suggest you consult our concept sheet on tourism and its consequences in order to determine if some of them can be applied to the Maritime Greenwich : https://www.alloprof.qc.ca/fr/eleves/bv/ geographie/tourisme-definitions-histoire-et-impacts-g1024 I hope that it will help you in your research. Do not hesitate to write to us again if you have any other questions! :) \\
  \hline
  \end{tabular}
  \caption{An example of a typical question and explanation on the Alloprof website with a relevant link in the answer. Questions have an average word count of 31 words and explanations 100 words.}
  \label{tab:example}
\end{table}

\subsection{Characteristics}
\label{sec:characteristics}

\begin{figure}[ht]
  \includegraphics[width=.5\textwidth]{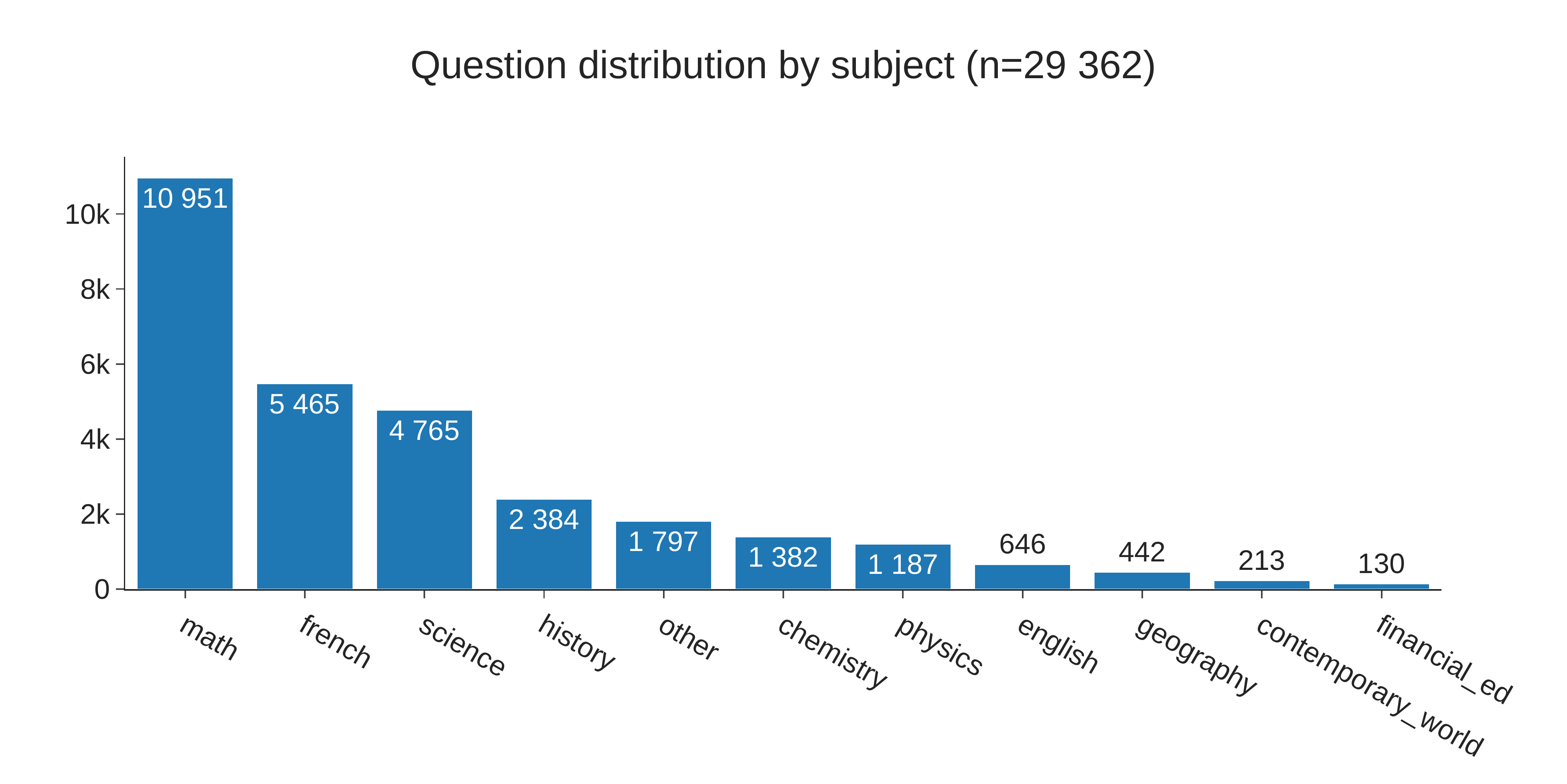}
  \caption{The majority of the questions are in mathematics~(37.3\%), French~(18.6\%) and science~(16.2\%), while all other subjects cover the remaining 27.9\% of questions.}
  \label{fig:dist-subjects}
\end{figure}

\begin{figure}[ht]
  \includegraphics[width=.5\textwidth]{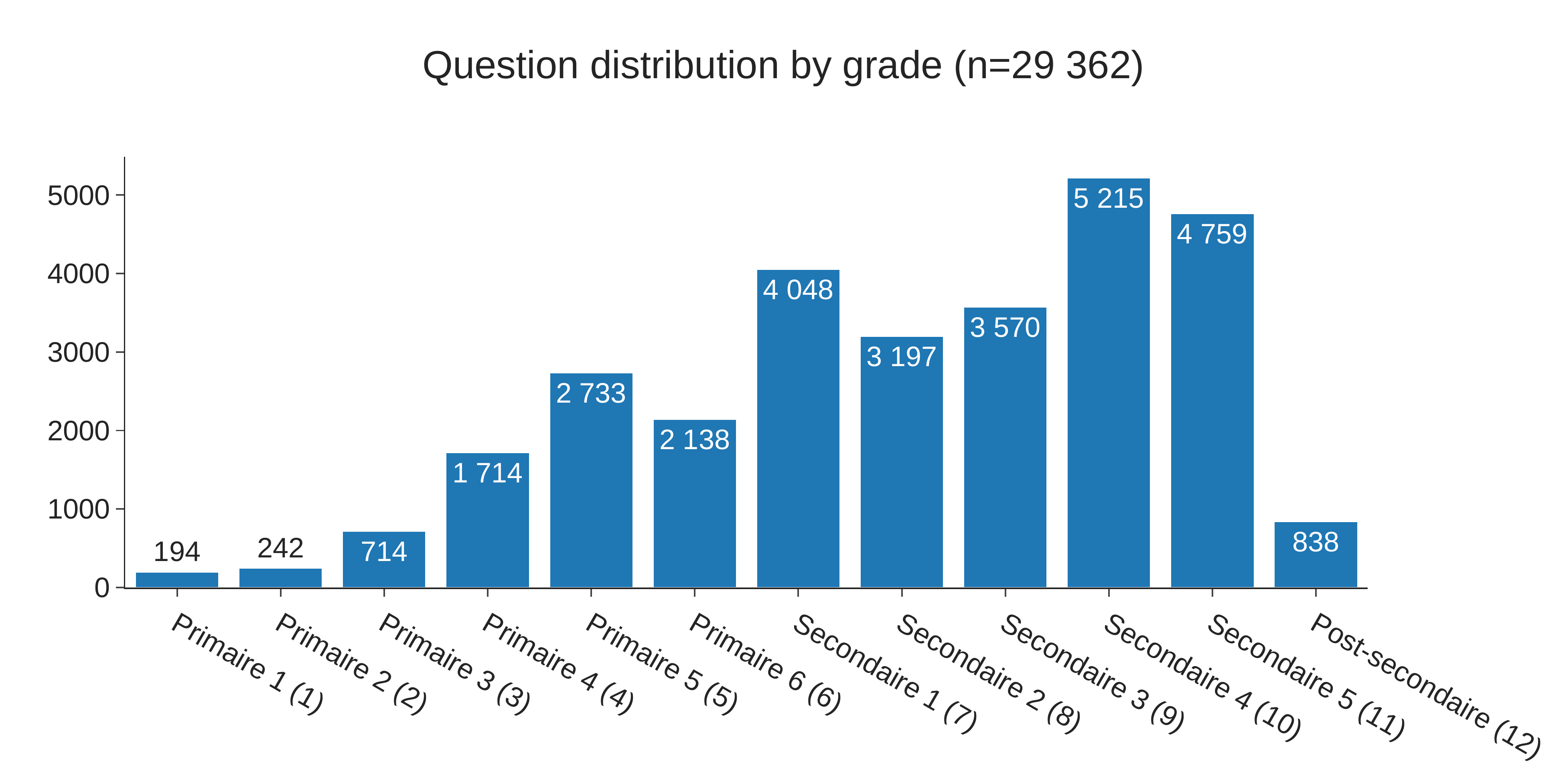}
  \caption{While most questions are high school level (\textit{secondaire}), the differences between grades, at least from grade 4 upwards, are not as important as the difference between subjects. Equivalent K12 grades are shown within parenthesis.}
  \label{fig:dist-grades}
\end{figure}

With some exceptions, questions are in French (97.4\%) with the rest in English, but reference pages are available in both languages, and figures~\ref{fig:dist-subjects} and \ref{fig:dist-grades} show us that the distribution of questions by subject and grade is far from uniform.

Both reference pages and questions can differ greatly in their makeup depending on the subject, with documents in the STEM subjects often having a lot of formulas and numbers compared to documents in other subjects.

Questions, and to a lesser extent reference pages, also differ considerably between grade levels with respect to the complexity of the language and the number of spelling mistakes.

Another characteristic of the dataset is the presence of images in a quarter of the questions (24.7\%), often containing text that solely refers to the image or elements of the image.
For example, \textit{[IMG] can you give me the examples for each situations? because i don't get them. thanks}.

\begin{figure}[ht]
  \includegraphics[width=.5\textwidth]{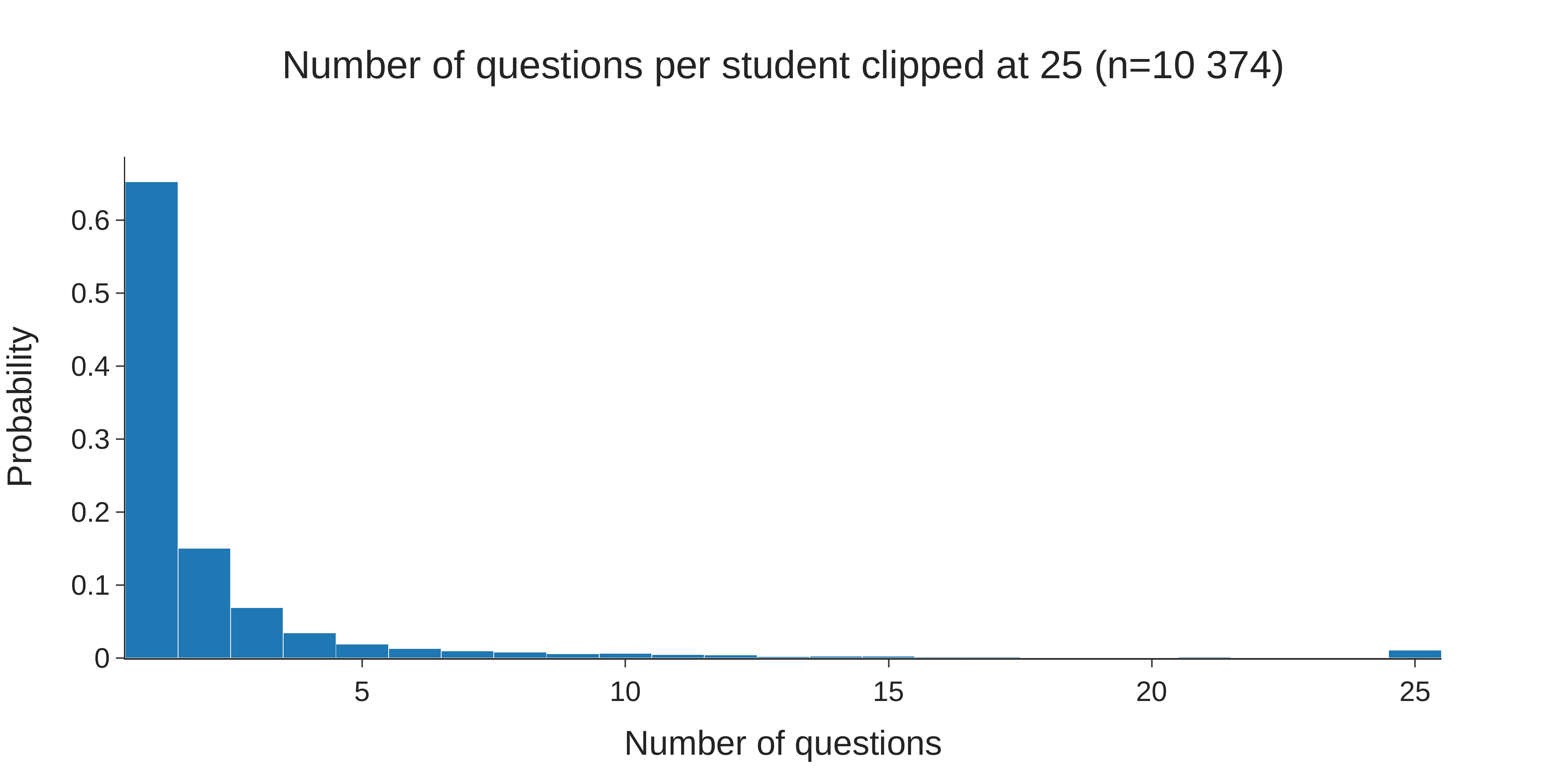}
  \caption{Most students have asked a single question, but following a power law distribution, around a third have asked more. Here the number of asked questions has been clipped at 25 to better show the distribution at lower numbers.}
  \label{fig:dist-per-user}
\end{figure}

Most students (65.3\%) have only asked one question, but figure~\ref{fig:dist-per-user} shows that a good number have asked more, following a power law distribution.

\subsection{Information Retrieval}
\label{sec:ir}

While this dataset is composed of questions and explanations, we explored and adapted it for an information retrieval task in the educational setting where the query is the question.
For a given question, the information retrieval system's goal is to find a limited number of reference pages on the site or similar questions asked in the past.
In an ideal setting for training models, we would know all the relevant documents (question or reference page) for a given question and could assume others are not.
Unfortunately, we do not have this information and so a proxy variable must be used.

This proxy variable is based on links provided in accepted answers (see figure~\ref{fig:alloprof}).
For around half the questions in the dataset (55.1\%), the explanation has at least one link either to another question or to a reference page on the website and these are the ones considered relevant to the question.
Because other documents can also be relevant, appropriate metrics and the ones we present in section~\ref{sec:evaluation} should focus on the presence of these linked documents and not their absence.

\section{Case study on relevant document retrieval task}
\label{sec:case-study}

As mentioned previously, a second contribution of this paper is to provide baseline performances on a document retrieval task for this dataset, returning the most relevant questions or reference pages (which we call documents) using the proxy variable of a mentioned link both for training and evaluating models.
To make these predictions, we followed the recent literature on other datasets and used Transformer based architectures~\cite{guo20,tra21}.

\begin{figure}[ht]
  \includegraphics[width=.5\textwidth]{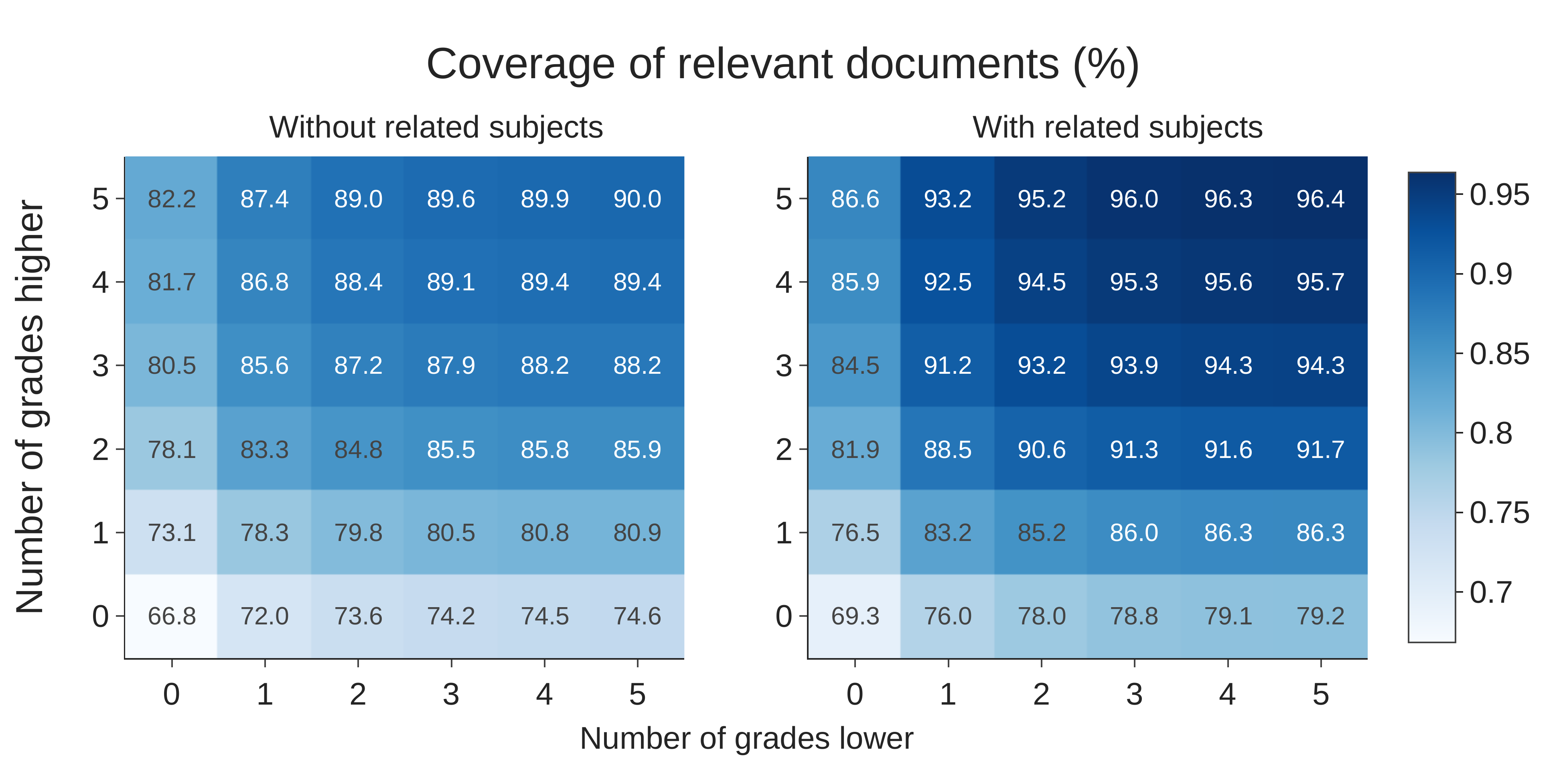}
  \caption{The proportion of relevant documents in the list of possible ones (coverage) depends highly on allowing related subjects in that list of possible documents as well as allowing document grades to be higher or lower than the question grade.}
  \label{fig:coverage}
\end{figure}

To limit the number of documents to evaluate and rank for a given question, a set of possible documents must be created, which is the limited set of documents from which a relevant one can be suggested for the question by the model.
The most trivial way of doing this is to limit these possible documents to ones of the same subject and grade level.
Since each reference page is labeled as relevant to many grades, this approach covers 66.8\% of the links given in explanations, limiting the choice of the model too much compared to what humans recommend in practice.
Figure~\ref{fig:coverage} shows the proportion of these links covered by the set of possible documents created when relaxing the number of lower and higher grade levels as well as allowing related subjects to be part of that set of possible documents (e.g. science for physics, geography for history, etc.).
To keep as much of the relevant links as possible, while also limiting the number of documents the model must rank, related subjects and $\pm 5$ grades were used to create the set of possible documents for training and evaluating.
In the final testing, we evaluated the impact of using different combinations of possible documents on both the evaluation metrics and the inference time~(see section~\ref{sec:results}).

The problem formulation and the way data is structured will be different when training compared to testing or inference.
The way the model is used when testing, reflecting how it would be used in a production setting, is to evaluate the relevance between a given question and all possible documents and choose the top $k$ best documents.
While in theory, this $k$ is arbitrary, results presented here are based on three recommendations following what is currently used on the site. 

But when training, the problem is structured as a binary pointwise task, predicting if a given question and document combination should be relevant or not.
When creating training minibatches, a number of relevant pairs are chosen and then other pairs are created by sampling a random non-relevant document from the list of possible ones for each question.
Following the work of other researchers~\cite{hof21,nug19}, we also tried different proportions and strategies than this one-to-one ratio, but it had little effect in practice and therefore only report results for this simpler approach.

\subsection{Evaluation}
\label{sec:evaluation}

As mentioned in section~\ref{sec:ir}, a good metric must evaluate the presence or absence of a good recommendation, and not the presence or absence of a bad one.
To measure and compare the model performances, we chose two common information retrieval metrics, Mean Reciprocal Rank (MRR) and normalised Discounted Cumulative Gain (nDCG), to which we added another simpler one we called \texttt{has\_correct} that corresponds to the way the model will be used and is the main metric we compare models with.
All three metrics will have results between 0 and 1 with a higher score being better.

For an ordered list of recommended documents for a question, the reciprocal rank is the inverse of the rank of the first document that is relevant.
For example, if the first recommendation is relevant, this will be 1.
If it is not and the second is, it will be 0.5 ($\frac{1}{2}$).
If the first two are not relevant and the third is, it will be $0.\overline{3}$ ($\frac{1}{3}$), and so on.
The mean reciprocal rank will simply be the mean of this reciprocal rank over all questions.

While MRR depends only on the first predicted relevant document, nDCG takes into account the rank of all relevant documents.
The formula to calculate it is:

\begin{equation}
  \begin{split}
    nDCG_q & = \frac{DCG_q}{IDCG_q} \\
    DCG_q & = \sum_{i=1}^{|rel_q|}\frac{1}{\log_2(r_{qi} + 1)} \\
    IDCG_q & = \sum_{i=1}^{|rel_q|}\frac{1}{\log_2(i + 1)} \\
    nDCG & = \frac{1}{|Q|}\sum_{q \in Q}nDCG_q \\
    rel_q & := \text{set of relevant documents for question }q \\
    r_{qi} & := \text{rank of the }i\text{th relevant document for question }q
  \end{split}
\end{equation}

For the \texttt{has\_correct} metric, it simply is the proportion of questions for which the recommended document list has at least one relevant document.

For all training and model comparisons, we used the related subjects and $\pm 5$ grades when creating the lists of possible documents to make sure we kept as many relevant documents as possible.
For final testing, we evaluated the impact of using these different ways of determining the possible documents on both the evaluation metrics and the inference time (see section~\ref{sec:results}).

\subsection{Models}

We tested two varieties of transformer models for which the trade-off is mainly around predictive performance versus inference time.

A transformer model is a type of neural network that is well suited to finding meaning in sequential data (such as text) by finding relationships between different elements of the sequence (e.g. words) using a self-attention mechanism~\cite{vas17}.
One commonly used model based on a transformer architecture for representing text is BERT~\cite{dev19} which was pre-trained on a variety of different datasets by trying to predict the right words that were masked (hidden from the model).
This allowed the model to learn the semantic and syntactic meanings of different words in different contexts, allowing it to comprehend the meaning of whole texts.
Because the datasets it was trained on were most often collected from the Internet, the model was exposed to misspellings, bad grammar and a variety of ways of writing.
Both of the models we present here use BERT as part of their architecture.

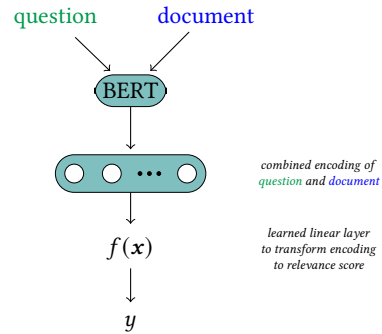
\begin{figure}[ht]
  \begin{center}
    \begin{tikzpicture}[
  dot/.style = {circle, fill, minimum size=#1,
  inner sep=0pt, outer sep=0pt},
  dot/.default = 6pt
  ]  

  \node[anchor=base, text=green] (q) at (0, 0) {question};
  \node[anchor=base, text=blue] (d) at (2, 0) {document};

  \node[
    rectangle,
    very thin,
    fill=teal!50,
    rounded corners=.25cm,
    draw,
    anchor=base
    ] (b) at (1, -1) {BERT};

  \draw[->, shorten >=2pt] (q) -- (b);
  \draw[->, shorten >=2pt] (d) -- (b);

  \node[
    rectangle,
    very thin,
    fill=teal!50,
    rounded corners=0.25cm,
    minimum height=0.5cm,
    minimum width=2cm,
    draw
    ] (e) at (1, -2) {};
  \foreach \x in {1, 2, 4} {
    \node[
      dot=0.25cm,
      very thin,
      fill=white,
      draw
      ] at (-0.25 + \x/2, -2) {};
  }
  \foreach \x in {-1, 0, 1} {
    \node[
      dot=0.05cm,
      very thin,
      fill=black,
      draw
      ] at (1.25 + \x * 0.125, -2) {};
  }

  \draw[->, shorten >=2pt] (b) -- (e);

  \node[align=center, font=\tiny] at (3.5, -2) {
      \textit{combined encoding of} \\
      \textit{\textcolor{green}{question} and \textcolor{blue}{document}}
  };

  \node (f) at (1, -3) {$f(\vec{x})$};
  \draw[->, shorten >=2pt] (e) -- (f);
  \node[align=center, font=\tiny] at (3.5, -3) {
    \textit{learned linear layer} \\
    \textit{to transform encoding} \\
    \textit{to relevance score}
  };

  \node (y) at (1, -4) {$y$};
  \draw[->, shorten >=2pt] (f) -- (y);

\end{tikzpicture}
  \end{center}
  \caption{The \texttt{TransformerCat} model encoded the question and document together before using the representation to directly predict the relevance between the two.}
  \label{fig:transformercat}
\end{figure}

The first one, \texttt{TransformerCat}, is an interaction-based system that concatenates a question and given document before passing it to a BERT model pre-trained on various corpora in French or multi-lingual~\cite{con19,le20,mar20}.
Because the question and document are encoded together by the transformer, it can pay attention to the words of the question based on the words in the document (and vice-versa) and directly compute through a function the relevance between the two (see figure~\ref{fig:transformercat}).

\begin{figure}[ht]
  \begin{center}
    \begin{tikzpicture}[
  dot/.style = {circle, fill, minimum size=#1,
  inner sep=0pt, outer sep=0pt},
  dot/.default = 6pt
  ]  

  \node[anchor=base, text=green] (q) at (0, 0) {question};
  \node[anchor=base, text=blue] (d) at (2, 0) {document};

  \node[
    rectangle,
    very thin,
    fill=green!50,
    rounded corners=.25cm,
    draw,
    anchor=base
    ] (bq) at (0, -1) {BERT};
  \node[
    rectangle,
    very thin,
    fill=blue!50,
    rounded corners=.25cm,
    draw,
    anchor=base
    ] (bd) at (2, -1) {BERT};

  \draw[->, shorten >=2pt] (q) -- (bq);
  \draw[->, shorten >=2pt] (d) -- (bd);

  \node[
    rectangle,
    very thin,
    fill=green!50,
    rounded corners=0.25cm,
    minimum height=0.5cm,
    minimum width=2cm,
    draw
    ] (eq) at (0, -2) {};
  \foreach \x in {1, 2, 4} {
    \node[
      dot=0.25cm,
      very thin,
      fill=white,
      draw
      ] at (-1.25 + \x/2, -2) {};
  }
  \foreach \x in {-1, 0, 1} {
    \node[
      dot=0.05cm,
      very thin,
      fill=black,
      draw
      ] at (0.25 + \x * 0.125, -2) {};
  }

  \node[
    rectangle,
    very thin,
    fill=blue!50,
    rounded corners=0.25cm,
    minimum height=0.5cm,
    minimum width=2cm,
    draw
    ] (ed) at (2, -2) {};
  \foreach \x in {1, 2, 4} {
    \node[
      dot=0.25cm,
      very thin,
      fill=white,
      draw
      ] at (0.75 + \x/2, -2) {};
  }
  \foreach \x in {-1, 0, 1} {
    \node[
      dot=0.05cm,
      very thin,
      fill=black,
      draw
      ] at (2.25 + \x * 0.125, -2) {};
  }

  \draw[->, shorten >=2pt] (bq) -- (eq);
  \draw[->, shorten >=2pt] (bd) -- (ed);

  \node[align=center, font=\tiny] at (4.5, -2) {
      \textit{separate encoding of} \\
      \textit{\textcolor{green}{question} and \textcolor{blue}{document}}
  };

  \node (f) at (1, -3) {$f(\vec{x})$};
  \draw[->, shorten >=2pt] (eq) -- (f);
  \draw[->, shorten >=2pt] (ed) -- (f);
  \node[align=center, font=\tiny] at (4.5, -3) {
    \textit{function to compute similarity} \\
    \textit{between question and document} \\
    \textit{encoding}
  };

  \node (y) at (1, -4) {$y$};
  \draw[->, shorten >=2pt] (f) -- (y);

\end{tikzpicture}
  \end{center}
  \caption{The \texttt{TransformerSim} model encodes the question and document separately and then compares the two with a function to get a similarity score.}
  \label{fig:transformersim}
\end{figure}
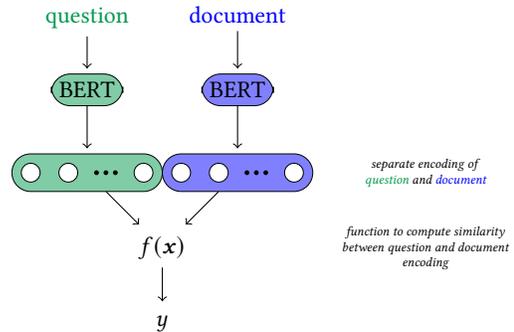

By contrast, the other model, \texttt{TransformerSim}, is a representation-based system.
It will also use a BERT model pre-trained on the same corpora as for \texttt{TransformerCat}, but will encode separately the question and the document before combining them with a function to get a similarity score which is interpreted as a relevance score in a similar way to siamese networks~(see figure~\ref{fig:transformersim})~\cite{rei19}.
This function can be any similarity function or a learned layer, but we found the dot product to work best by a large margin.
With this model, it is possible to fine-tune the same BERT encoder for both the question and the document, or to fine-tune different encoders for each of them, which we found to work better at the cost of doubling the number of model parameters.
The advantage of this solution is the possibility of encoding all documents beforehand and therefore only have to look up the wanted encodings at inference time, drastically speeding up the process.

For both models, the same pre-trained encoders were tried from~\cite{dev19,con19,le20,mar20}, which trained BERT models on either French corpora~\cite{le20,mar20} or multi-lingual corpora~\cite{dev19,con19}.
In our experiments, the \textit{camembert} family of encoders~\cite{mar20} in two sizes, \texttt{camembert-base} (110M parameters) and \texttt{camembert-large} (335M parameters), almost always performed better and so only their results are shown here.

Both types of models and all pre-trained encoders were fine-tuned on the dataset, using the Adam optimizer~\cite{kin15} with initial learning rate of $10^{-6}$ and early stopping.
In practice, training converged between 5 and 20 epochs for all combinations.

\subsection{Results}
\label{sec:results}

We show in table~\ref{tab:results} the best results for both models using the \textit{camembert}~\cite{mar20} family of pre-trained encoders, which had the best performances compared to other pre-trained models.
We see that in all cases, half or more of the questions had a top-3 predictions containing at least one relevant document, with only a 14\% difference between the best model, \texttt{TransformerCat} with \texttt{camembert-large}, and the worst one, \texttt{TransformerSim} with \texttt{camembert-base} on the \texttt{has\_good} metric.
But for the information retrieval system to be useful in practice, it has to be able to return a prediction in a timely fashion.
Optimizing the \texttt{TransformerSim} models to precompute the document encodings beforehand, the worst model based on prediction scores becomes almost 17 times faster.

\begin{figure}[ht]
  \includegraphics[width=.5\textwidth]{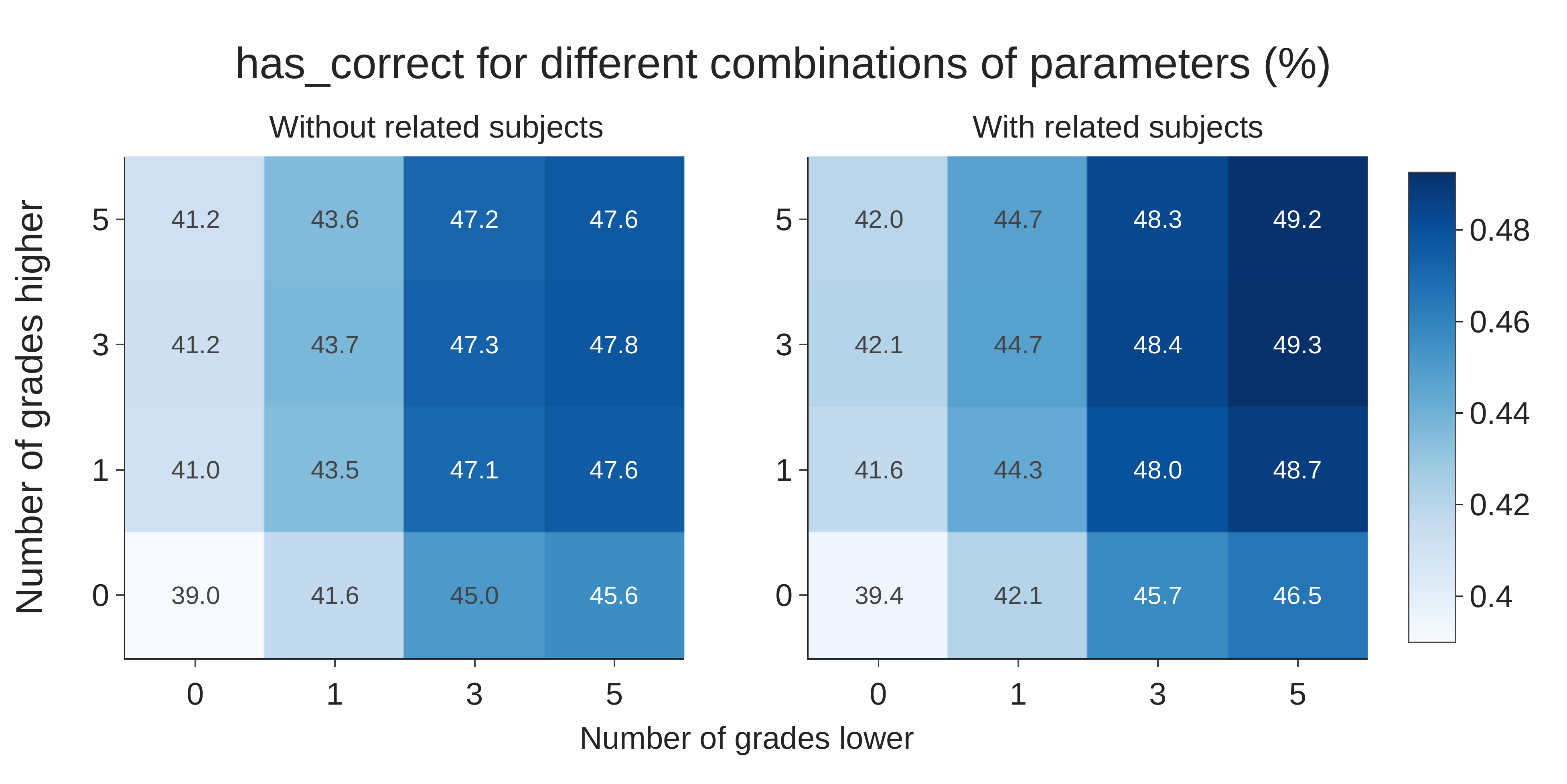}
  \caption{The prediction score is highly influenced by the inclusion of related subjects and the size of the lower and higher grade window~(see figure~\ref{fig:coverage}), which creates a difference of 10.2\% between the lowest coverage and the highest one.}
  \label{fig:result}
\end{figure}

As shown in figure~\ref{fig:coverage}, there is a large difference in the number of correct predictions that are possible across the inferior and superior number of grades, and whether related subjects are included or not.
This is reflected in figure~\ref{fig:result} where we see a 10\% difference in the result between the combination of parameters leading to the highest coverage and the combination leading to the lowest one.
While this is a large difference, it is much smaller than the 30\% difference in coverage, which tells us the model tends to give recommendations closer to the same grade level and of the same subject than what humans would recommend.
Whether that is a good or bad thing is a matter for discussion between subject matter experts outside the scope of this paper.

\begin{figure}[ht]
  \includegraphics[width=.5\textwidth]{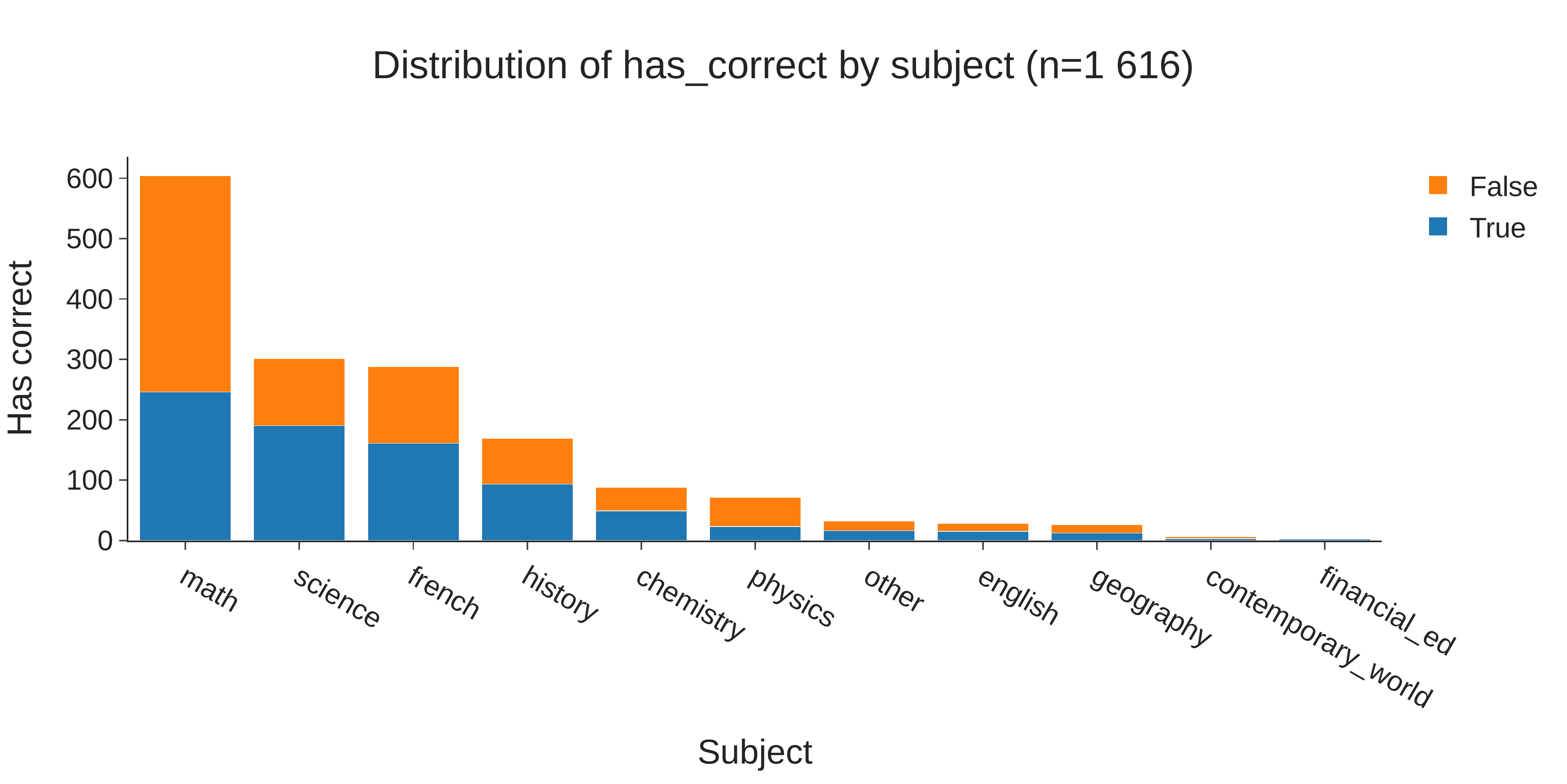}
  \caption{Results vary considerably by subject. Mathematics is by far the one with the most mistakes (and the most common one). Results here are based on including related subjects and using 5 lower and higher grades.}
  \label{fig:result-subject}
\end{figure}

We also see in figure~\ref{fig:result-subject} that the success of the model varies considerably between subjects, getting results well above 50\% on all subjects except for math and physics.
Those being the two subjects with questions most likely containing critical information in symbols and images rather than natural text, it makes sense that a model pre-trained on mostly open text would not perform as well.

Most explanations contain a single link and so \texttt{has\_correct} is the same as recall for those associated questions, but in table~\ref{tab:recall} we see that there is a big drop in recall performance when more than one link is present.
We hypothesize, but have not verified other than qualitatively through a few examples, that the reason for this is that most often there is one link to a reference page closely aligned to the question and the others are there to give more context to the explanation and provide further readings of interest that are not directly related to the question.

Other experiments were considered to try improving results such as correcting the language of questions beforehand and encoding formulas with a different pre-trained model, but none of them improved the baseline model and so are not reported here.

As specified in an earlier section, the questions and reference pages also contain images, and for many questions all the information is contained in these images.
Our only crude experiment using Tesseract OCR~\cite{smi07} to extract the text from the images led to worse results, often transcribing only part of the characters and creating gibberish text.
A qualitative analysis of the predictions, especially in math, the most common subject, showed that many of the wrong predictions were for those with images and so further research in extracting text or information is worth exploring to potentially greatly improve the current results.

\begin{table*}[ht]
  \begin{center}
    \begin{tabular}{c c c c c c c c}
      & \textbf{Model} & \textbf{Pre-trained encoder} & \textbf{\texttt{has\_good\_3}} & \textbf{\texttt{mrr}} & \textbf{\texttt{ndcg}} & \textbf{Inference time (ms)} & \textbf{Size (G)} \\
      \hline
      Previous model & TF-IDF & - & 0.242 & 0.203 & 0.316 & - & - \\
      \hline
      \multirow{4}{*}{Our models} & \multirow{2}{*}{\texttt{TransformerCat}} & \texttt{camembert-large} & \textbf{0.585} & \textbf{0.537} & \textbf{0.618} & 12812 & 3.8 \\
                                  & & \texttt{camembert-base} & 0.562 & 0.512 & 0.596 & 5354 & \textbf{1.3} \\
                                  & \multirow{2}{*}{\texttt{TransformerSim}} & \texttt{camembert-large} & 0.522 & 0.475 & 0.568 & 1428 & 7.6 \\
                                  & & \texttt{camembert-base} & 0.512 & 0.467 & 0.560 & \textbf{720} & 2.5
    \end{tabular}
  \end{center}
  \caption{The best transformer model is \texttt{TransformerCat} which is able to find more complex patterns between the query and document texts at the cost of needing much more computation at inference time. And while the bigger \texttt{camembert-large} encoders perform better than their smaller counterpart, it is not a big improvement, especially when considering the doubling of inference time and almost three times larger model size. We show performances for the previously used model based on TF-IDF and cosine similarity for comparison.}
  \label{tab:results}
\end{table*}

\begin{table}[ht]
    \begin{tabular}{l r r}
      \textbf{\# relevant} & \textbf{\# questions} & \textbf{Recall} \\
      \hline
      1 & 1323 & 0.525 \\
      2 & 209 & 0.380 \\
      3 & 55 & 0.345 \\
      4+ & 29 & 0.221 \\
    \end{tabular}
  \caption{Most questions~(82.9\%) only have an explanation with a single link and half of those are correctly predicted. When there is more than one link, the model has trouble predicting the extra links, potentially indicating that those are only tangentially  linked to the query and that the model only gives straightforward recommendations.}
  \label{tab:recall}
\end{table}

\begin{figure}[ht]
  \includegraphics[width=.5\textwidth]{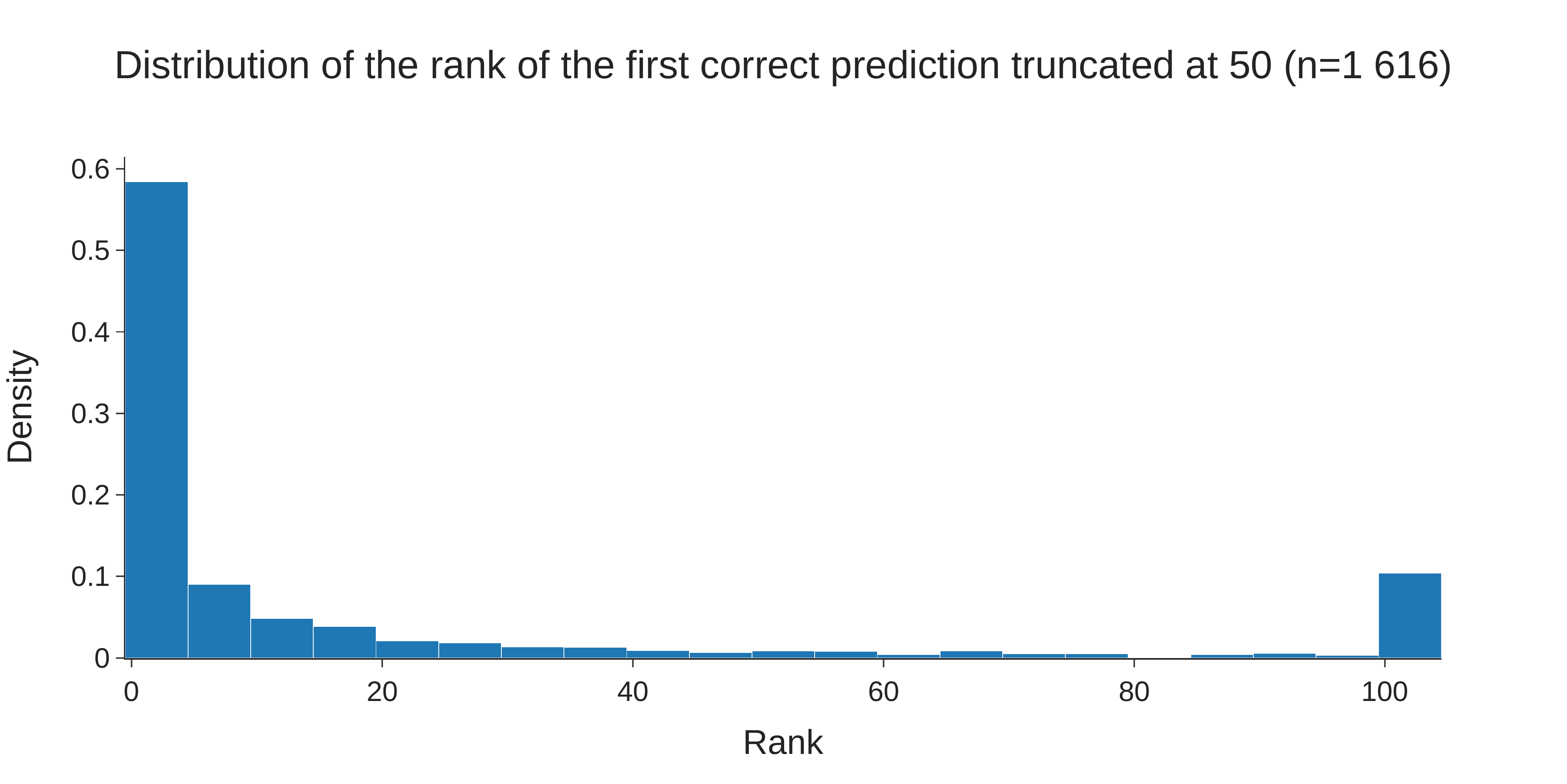}
  \caption{For most questions, there is at least one relevant document in the top predictions. But the rest are dispersed randomly in other ranks, indicating that for those, the model gleans no understanding of the relationship between the query and relevant documents.}
  \label{fig:ranks}
\end{figure}

\begin{figure}[ht]
  \includegraphics[width=.5\textwidth]{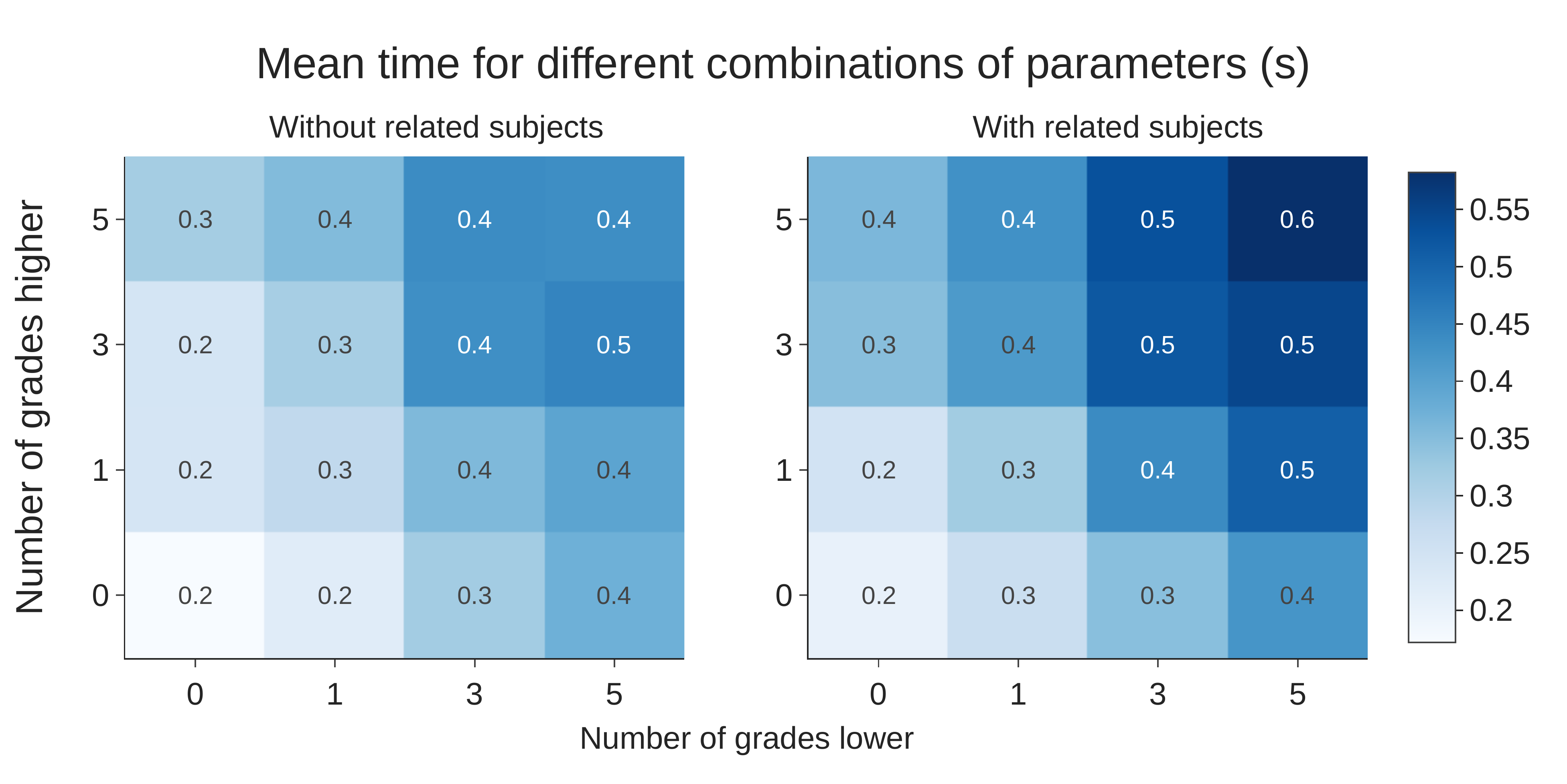}
  \caption{The trade-off in adding related subjects and bigger grade level windows is an increase in inference time, tripling between the lower and upper ends.}
  \label{fig:time-mean}
\end{figure}

\begin{figure}[ht]
  \includegraphics[width=.5\textwidth]{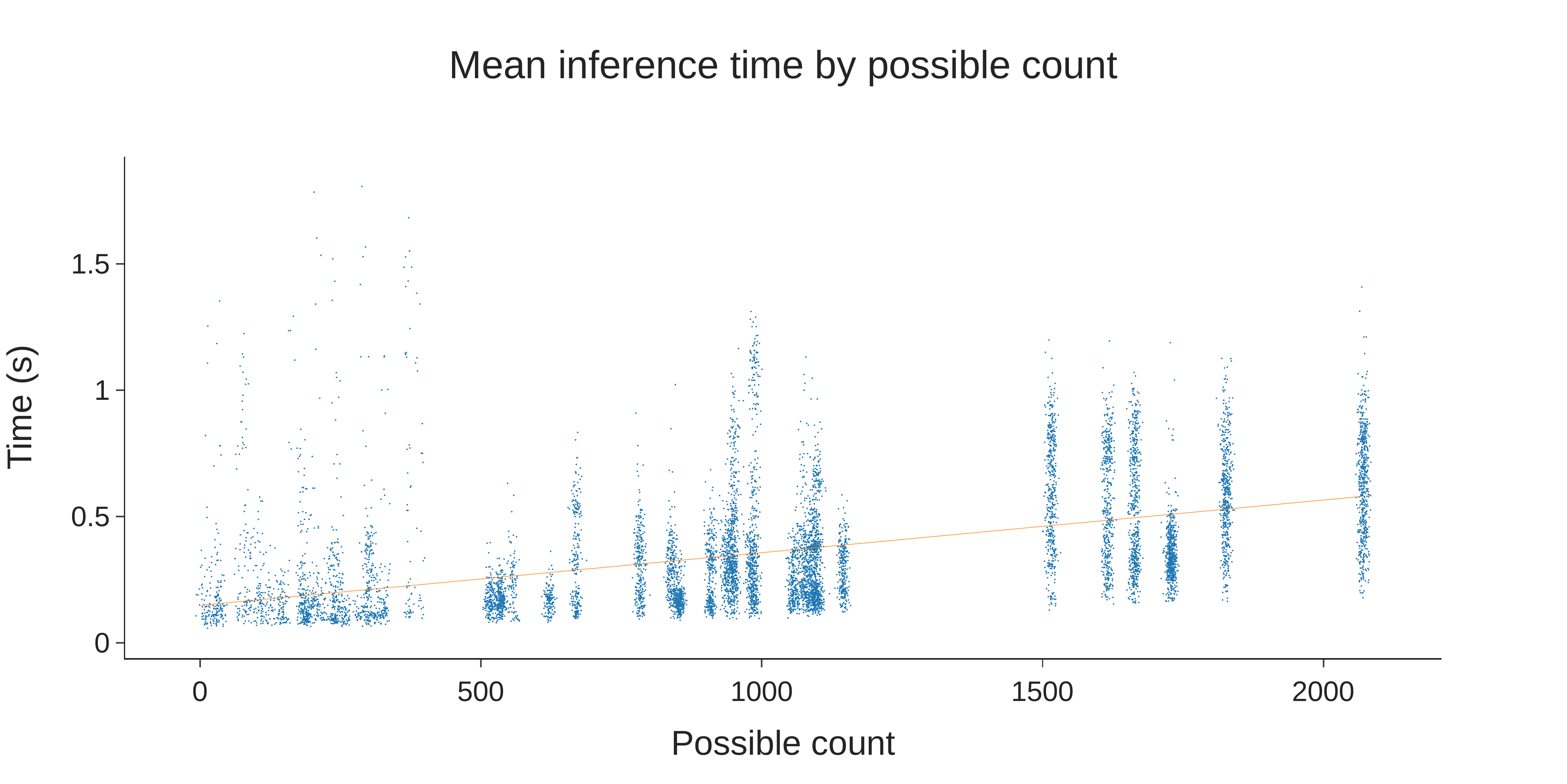}
  \caption{Inference time increases as a function of the number of possible documents from which to recommend.}
  \label{fig:time-possible}
\end{figure}

While the correctness of predictions was the focus of most of the work, as important to making this system a useful one was inference time on CPU.
Alloprof is a non-profit organisation and as such needs to limit costs to be able to help as much as possible, making the use of GPUs at inference impossible, and needs to be quick as it will be used by children in a context where it is easy for them to get distracted or frustrated.
The other trade-off is that, as seen in figures~\ref{fig:coverage} and \ref{fig:result}, the more possible documents there are to recommend, the more the model has a chance to find the exact right one.
Figures~\ref{fig:time-mean} and \ref{fig:time-possible} show that the inference time increases linearly with the number of possible documents to predict and while the mean inference time stays reasonable in all our experiments, it can become an issue in the upper limits of the distribution.

\section{Conclusion}

Students and teachers rely more and more on the continually improving learning material found online.
Alloprof, a non-profit based in Quebec, offers such high-quality material on its website as well as a forum where students may ask questions that will eventually be answered or validated by teachers.
For many of these questions, the explanation can be found on one of the reference pages written by Alloprof, but it can take hours for a teacher to eventually give that link to the student and many questions are too complex for a search engine to parse and find the right resource.

We introduce in this paper a new and open dataset built from 29~349 of these questions and explanations collected over six months from 10~368 students, as well as the 2~596 reference pages.
55.1\% of these explanations contain at least one link either to another question or a reference page, allowing the dataset to be easily adapted to an information retrieval task.
The students who asked these questions are from primary and secondary schools in Quebec, corresponding roughly to K12 students in the American system, leading to a variety of thought complexity and language mistakes in their text.
Students are also allowed to use images in their questions, 24.7\% of the questions have an image in them ranging from a clear picture of a printed question text to a blurry picture of their handwritten answer to an exam question, through various mathematical diagrams or tables.
This reflects the increasing tendency of students growing up with smartphones to take advantage of the ease of taking a snapshot to give a better context of what they are trying to explain.
\footnote{Raw and transformed data is available at both \url{https://github.com/mila-aia/alloprof-data} and \url{https://huggingface.co/datasets/antoinelb7/alloprof}}

We also present a case study of the use of this dataset in an information retrieval task, where the goal is to give a student the corresponding reference page or another similar and already asked question in real-time using a transformer architecture.
Fine-tuning pre-trained French encoders and recommending three documents (reference page or question), our models obtained at least one document that would eventually be recommended by a teacher in their explanation at least 50\% of the time and up to 58.5\% of the time.
While one type of architecture had an inference time that was too slow to be used in real-time, the other one can offer predictions in well under one second using a CPU.

While the results are good enough for this baseline to be used in production, there is still a lot of future work to do in the many tasks this dataset can be applied to.
In the information retrieval one we explored, the treatment of images is necessary for the many questions for which most of the information is contained in the image and for which both our system and search engines are blind to.
Students asking questions often ask them after they found and looked at the available learning material, but have trouble understanding it or finding their exact answer in it.
Using this dataset to extract or generate the right explanation for a given question is not something we explored, but does lead to interesting research questions.
Can multiple reference pages be combined together to generate a better explanation?
Can a reference page or a subset of it be summarized or rewritten in a simpler language for younger students?
How can the explanation be structured to better guide the student towards the answer?
This last question also leads to the holy grail in the context of education and a possible use of the dataset: creating an iterative learning recommender system.
While 65.3\% of students have only asked one question, is it possible to look at the other ones and use their question history in a recommender system to guide them in a learning path?

\printbibliography

@inproceedings{ngu16,
  title={MS MARCO: A human generated machine reading comprehension dataset},
  author={Nguyen, Tri and Rosenberg, Mir and Song, Xia and Gao, Jianfeng and Tiwary, Saurabh and Majumder, Rangan and Deng, Li},
  booktitle={CoCo@ NIPs},
  year={2016}
}

@inproceedings{sob21,
  title={Overview of TREC 2021},
  author={Soboroff, Ian},
  booktitle={30th Text REtrieval Conference. Gaithersburg, Maryland},
  year={2021}
}

@book{har93,
  title={The first text retrieval conference (TREC-1)},
  author={Harman, Donna K},
  volume={500},
  number={207},
  year={1993},
  publisher={US Department of Commerce, National Institute of Standards and Technology}
}

@inproceedings{lou22,
  title = {A Statutory Article Retrieval Dataset in French},
  author = {Louis, Antoine and Spanakis, Gerasimos},
  booktitle = {Proceedings of the 60th Annual Meeting of the Association for Computational Linguistics},
  month = may,
  year = {2022},
  address = {Dublin, Ireland},
  publisher = {Association for Computational Linguistics},
  url = {},
  doi = {},
  pages = {To appear},
}

@article{zha22,
  title={Making a MIRACL: Multilingual Information Retrieval Across a Continuum of Languages},
  author={Xinyu Zhang and Nandan Thakur and Odunayo Ogundepo and Ehsan Kamalloo and David Alfonso-Hermelo and Xiaoguang Li and Qun Liu and Mehdi Rezagholizadeh and Jimmy Lin},
  journal={ArXiv},
  year={2022},
  volume={abs/2210.09984}
}

@article{tra21,
  title={Neural ranking models for document retrieval},
  author={Trabelsi, Mohamed and Chen, Zhiyu and Davison, Brian D and Heflin, Jeff},
  journal={Information Retrieval Journal},
  volume={24},
  number={6},
  pages={400--444},
  year={2021},
  publisher={Springer}
}

@article{guo20,
  title={A deep look into neural ranking models for information retrieval},
  author={Guo, Jiafeng and Fan, Yixing and Pang, Liang and Yang, Liu and Ai, Qingyao and Zamani, Hamed and Wu, Chen and Croft, W Bruce and Cheng, Xueqi},
  journal={Information Processing \& Management},
  volume={57},
  number={6},
  pages={102067},
  year={2020},
  publisher={Elsevier}
}

@inproceedings{nug19,
  title={Strategy of the negative sampling for training retrieval-based dialogue systems},
  author={Nugmanova, Aigul and Smirnov, Andrei and Lavrentyeva, Galina and Chernykh, Irina},
  booktitle={2019 IEEE International Conference on Pervasive Computing and Communications Workshops (PerCom Workshops)},
  pages={844--848},
  year={2019},
  organization={IEEE}
}

@inproceedings{hof21,
  title={Efficiently teaching an effective dense retriever with balanced topic aware sampling},
  author={Hofst{\"a}tter, Sebastian and Lin, Sheng-Chieh and Yang, Jheng-Hong and Lin, Jimmy and Hanbury, Allan},
  booktitle={Proceedings of the 44th International ACM SIGIR Conference on Research and Development in Information Retrieval},
  pages={113--122},
  year={2021}
}

@inproceedings{dev19,
  title={BERT: Pre-training of Deep Bidirectional Transformers for Language Understanding},
  author={Devlin, Jacob and Chang, Ming-Wei and Lee, Kenton and Toutanova, Kristina},
  booktitle={Proceedings of NAACL-HLT},
  pages={4171--4186},
  year={2019}
}

@inproceedings{mar20,
  title={CamemBERT: a Tasty French Language Model},
  author={Martin, Louis and Muller, Benjamin and Su{\'a}rez, Pedro Javier Ortiz and Dupont, Yoann and Romary, Laurent and de la Clergerie, {\'E}ric Villemonte and Seddah, Djam{\'e} and Sagot, Beno{\^\i}t},
  booktitle={ACL 2020-58th Annual Meeting of the Association for Computational Linguistics},
  year={2020}
}

@inproceedings{le20,
  title={FlauBERT: Unsupervised Language Model Pre-training for French},
  author={Le, Hang and Vial, Lo{\"\i}c and Frej, Jibril and Segonne, Vincent and Coavoux, Maximin and Lecouteux, Benjamin and Allauzen, Alexandre and Crabb{\'e}, Benoit and Besacier, Laurent and Schwab, Didier},
  booktitle={Proceedings of the 12th Language Resources and Evaluation Conference},
  pages={2479--2490},
  year={2020}
}

@article{con19,
  title={Cross-lingual language model pretraining},
  author={Conneau, Alexis and Lample, Guillaume},
  journal={Advances in neural information processing systems},
  volume={32},
  year={2019}
}

@inproceedings{kin15,
  title={Adam: A Method for Stochastic Optimization},
  author={Kingma, Diederik P and Ba, Jimmy},
  booktitle={ICLR (Poster)},
  year={2015}
}

@inproceedings{smi07,
  title={An overview of the Tesseract OCR engine},
  author={Smith, Ray},
  booktitle={Ninth international conference on document analysis and recognition (ICDAR 2007)},
  volume={2},
  pages={629--633},
  year={2007},
  organization={IEEE}
}

@article{vas17,
  title={Attention is all you need},
  author={Vaswani, Ashish and Shazeer, Noam and Parmar, Niki and Uszkoreit, Jakob and Jones, Llion and Gomez, Aidan N and Kaiser, {\L}ukasz and Polosukhin, Illia},
  journal={Advances in neural information processing systems},
  volume={30},
  year={2017}
}

@inproceedings{hof20,
  title={FQuAD: French Question Answering Dataset},
  author={Martin d'Hoffschmidt and Maxime Vidal and Wacim Belblidia and Quentin Heinrich and Tom Brendl'e},
  booktitle={Findings},
  year={2020}
}

@inproceedings{ker20,
  title={Project PIAF: Building a Native French Question-Answering Dataset},
  author={Rachel Keraron and Guillaume Lancrenon and Mathilde Bras and Fr{\'e}d{\'e}ric Allary and Gilles Moyse and Thomas Scialom and Edmundo-Pavel Soriano-Morales and Jacopo Staiano},
  booktitle={LREC},
  year={2020}
}

@article{kad22,
  title={FrBMedQA: the first French biomedical question answering dataset},
  author={Zakaria Kaddari and Toumi Bouchentouf},
  journal={IAES International Journal of Artificial Intelligence (IJ-AI)},
  year={2022}
}

@inproceedings{hei22,
  title={FQuAD2.0: French Question Answering and Learning When You Don’t Know},
  author={Quentin Heinrich and Gautier Viaud and Wacim Belblidia},
  booktitle={LREC},
  year={2022}
}

@article{red18,
  title={CoQA: A Conversational Question Answering Challenge},
  author={Siva Reddy and Danqi Chen and Christopher D. Manning},
  journal={Transactions of the Association for Computational Linguistics},
  year={2018},
  volume={7},
  pages={249-266}
}

@inproceedings{raj16,
  title={SQuAD: 100,000+ Questions for Machine Comprehension of Text},
  author={Pranav Rajpurkar and Jian Zhang and Konstantin Lopyrev and Percy Liang},
  booktitle={Conference on Empirical Methods in Natural Language Processing},
  year={2016}
}

@inproceedings{raj18,
  title={Know What You Don’t Know: Unanswerable Questions for SQuAD},
  author={Pranav Rajpurkar and Robin Jia and Percy Liang},
  booktitle={Annual Meeting of the Association for Computational Linguistics},
  year={2018}
}

@inproceedings{lai17,
  title={RACE: Large-scale ReAding Comprehension Dataset From Examinations},
  author={Guokun Lai and Qizhe Xie and Hanxiao Liu and Yiming Yang and Eduard H. Hovy},
  booktitle={Conference on Empirical Methods in Natural Language Processing},
  year={2017}
}

@article{abb20,
  title={Text‐based question answering from information retrieval and deep neural network perspectives: A survey},
  author={Zahra Abbasiyantaeb and Saeedeh Momtazi},
  journal={Wiley Interdisciplinary Reviews: Data Mining and Knowledge Discovery},
  year={2020},
  volume={11}
}

@article{nog19,
  title={Multi-Stage Document Ranking with BERT},
  author={Rodrigo Nogueira and Wei Yang and Kyunghyun Cho and Jimmy J. Lin},
  journal={ArXiv},
  year={2019},
  volume={abs/1910.14424}
}

@article{rob09,
  title={The Probabilistic Relevance Framework: BM25 and Beyond},
  author={Stephen E. Robertson and Hugo Zaragoza},
  journal={Found. Trends Inf. Retr.},
  year={2009},
  volume={3},
  pages={333-389}
}

@article{rei19,
  title={Sentence-BERT: Sentence Embeddings using Siamese BERT-Networks},
  author={Nils Reimers and Iryna Gurevych},
  journal={ArXiv},
  year={2019},
  volume={abs/1908.10084}
}

\end{document}